\documentclass[10pt,twocolumn,letterpaper]{article}
\usepackage[pagenumbers]{cvpr} 
\usepackage{url}
\usepackage{times}
\usepackage{epsfig}
\usepackage{graphicx}
\usepackage{amsmath}
\usepackage{amssymb}
\usepackage{multirow}
\usepackage[accsupp]{axessibility}
\usepackage{threeparttable}
\usepackage{booktabs}
\usepackage{setspace}
\usepackage{color}
\usepackage{algorithm}
\usepackage{algpseudocode}
\usepackage{algorithmicx}
\usepackage{marvosym}
\usepackage{amsmath}

\usepackage[pagebackref=true,breaklinks=true,colorlinks,bookmarks=false]{hyperref}
\usepackage[table,xcdraw]{xcolor}

\def\cvprPaperID{8215}
\def\confName{CVPR}
\def\confYear{2023}
\begin{document}
\title{Improving GAN Training via Feature Space Shrinkage }
\author{Haozhe Liu$^{1 \dagger}$, Wentian Zhang$^{2 \dagger}$,  Bing Li$^{1 \textrm{\Letter}}$, Haoqian Wu$^3$, Nanjun He$^2$, Yawen Huang$^2$ \\
Yuexiang Li$^{2  \textrm{\Letter}}$,Bernard Ghanem$^1$, Yefeng Zheng$^2$ \\
$^1$ AI Initiative, King Abdullah University of Science and Technology (KAUST), \\
$^2$Jarvis Lab, Tencent,  
$^3$YouTu Lab, Tencent\\
{\small \it \{haozhe.liu;bing.li;bernard.ghanem\}@kaust.edu.sa; zhangwentianml@gmail.com; }\\
{\small \it \{linuswu;yawenhuang;nanjunhe;vicyxli;yefengzheng\}@tencent.com} \\
}
\newcommand\blfootnote[1]{%
\begingroup
\renewcommand\thefootnote{}\footnote{#1}%
\addtocounter{footnote}{-1}%
\endgroup
}
\maketitle
\blfootnote{$\dagger$ Equal Contribution}
\blfootnote{Corresponding Authors: Bing Li and Yuexiang Li.}
\begin{abstract}
Due to the outstanding capability for data generation, Generative Adversarial Networks (GANs) have attracted considerable attention in unsupervised learning.
However, training GANs is difficult, since the training distribution is dynamic for the discriminator, leading to unstable image representation.
In this paper, we address the problem of training GANs from a novel perspective, \emph{i.e.,} robust image classification. Motivated by studies on robust image representation, we propose a simple yet effective module, namely AdaptiveMix, for GANs, which shrinks the regions of training data in the image representation space of the discriminator. 
Considering it is intractable to directly bound feature space, we propose to construct hard samples and narrow down the feature distance between hard and easy samples. The hard samples are constructed by mixing a pair of training images. We evaluate the effectiveness of our AdaptiveMix with widely-used and state-of-the-art GAN architectures. The evaluation results demonstrate that our AdaptiveMix can facilitate the training of GANs and effectively improve the image quality of generated samples. We also show that our AdaptiveMix can be further applied to image classification and Out-Of-Distribution (OOD) detection tasks, by equipping it with state-of-the-art methods. Extensive experiments on seven publicly available datasets show that our method effectively boosts the performance of baselines. 
The code is publicly available at \url{https://github.com/WentianZhang-ML/AdaptiveMix}. 

\end{abstract}

\section{Introduction}

\begin{figure}[!htbp]
    \centering
    \includegraphics[width=0.49\textwidth]{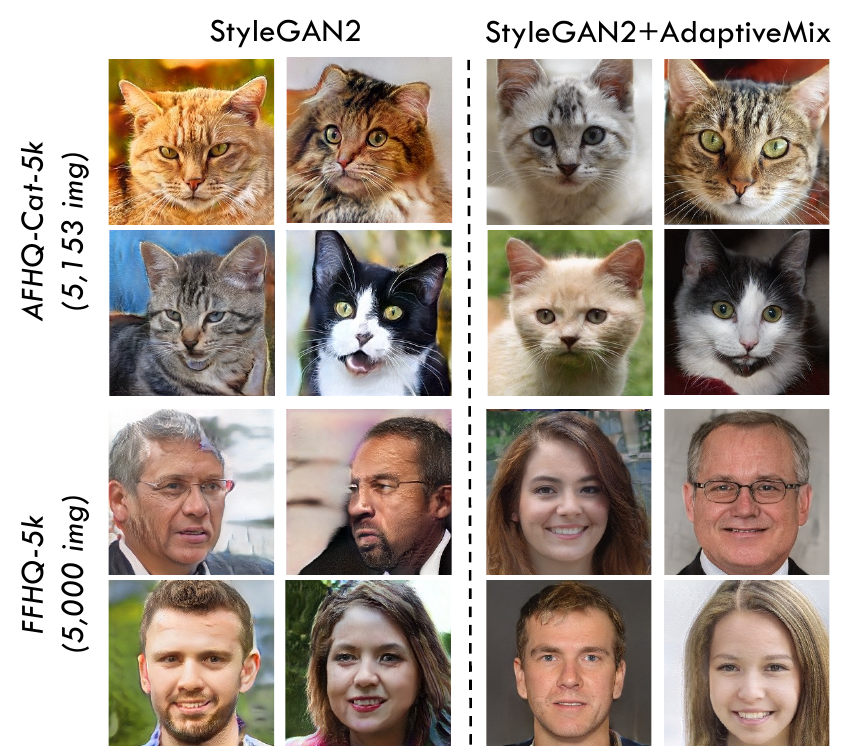}
    \caption{Results generated by StyleGAN-V2 \cite{karras2020analyzing} and our method (StyleGAN-V2 + AdaptiveMix) on AFHQ-Cat and FFHQ-5k. We propose a simple yet effective  module AdaptiveMix, which can be used for helping the training of unsupervised GANs.  When trained on a small amount of data,  StyleGAN-V2  generates images with artifacts, due to unstable training. However, our AdaptiveMix effectively boosts the performance of  StyleGAN-V2 in terms of image quality. }
   \vspace{-5pt}
    \label{fig:toys}
\end{figure}

Artificial Curiosity \cite{schmidhuber1991possibility,schmidhuber2020generative} and  Generative Adversarial Networks (GANs) have attracted extensive attention due to their remarkable performance in image generation \cite{karras2021alias,wang2018esrgan,zhang2019self,zhang2019ranksrgan}. A standard GAN consists of a generator and a discriminator network, where the discriminator is trained to discriminate real/generated samples, and the generator aims to generate samples that can fool the discriminator.
Nevertheless, the training of GANs is difficult and unstable, leading to low-quality generated samples \cite{miyato2018spectral,kurach2019large}. 

Many efforts have been devoted to improving the training of GANs (\eg \cite{goodfellow2014generative,salimans2016improved,heusel2017gans,miyato2018spectral,arjovsky2017wasserstein,liu2022combating,li2021anigan}). 
Previous studies \cite{radford2015unsupervised} attempted to co-design the network architecture of the generator and discriminator to balance the iterative training. Following this research line, PG-GAN \cite{karras2017progressive} gradually trains the GANs with progressive growing architecture according to the paradigm of curriculum learning. More recently, data augmentation-based methods, such as APA \cite{jiang2021deceive}, ADA \cite{karras2020training}, and adding noises into the generator \cite{karras2020analyzing}, were further proposed to stabilize the training of GANs.
A few works address this problem on the discriminator side. 
For example, WGAN \cite{arjovsky2017wasserstein} proposes to enforce a Lipschitz constraint by using weight clipping. Instead, WGAN-GP \cite{gulrajani2017improved} directly penalizes the norm of the discriminator's gradient. These methods have shown that revisions of discriminators can achieve promising performance. However, improving the training of GANs remains an unsolved and challenging problem.

In this paper, considering that the discriminator is critical to the training of GANs, we address the problem of training GANs from a novel perspective, \emph{i.e.,} robust image classification. 
In particular, the discriminator can be regarded as performing a classification task that  discriminates real/fake samples. Our insight is that controlling the image representation (\emph{i.e.,} feature extractor) of the discriminator can improve the training of GANs, motivated by studies on robust image classification \cite{verma2019manifold,liu2021group}.
More specifically, recent work \cite{verma2019manifold} on robust image representation presents  inspiring observations that training data is scattered in the learning space of vanilla classification networks; hence, the networks would improperly assign high confidences to samples that are off the underlying manifold of training data. This phenomenon also leads to the vulnerability of GANs, \emph{i.e.,} the discriminator cannot focus on learning the distribution of real data. Therefore, we propose to shrink the regions of training data in the image representation space of the discriminator. 

Different from existing works \cite{karras2020training,jiang2021deceive}, we explore a question for  GANs: \textit{Would the training stability of GANs be improved if we explicitly shrink the regions of training data in the image representation space supported by the discriminator?}
To this end, we propose a module named AdaptiveMix to shrink the regions of training data in the latent space constructed by a feature extractor. However, it is non-trivial and challenging to directly capture the boundaries of feature space. Instead, our insight is that
we can shrink the feature space by reducing the distance between hard and easy samples in the latent space, where hard samples are regarded as the samples that are difficult for classification
networks to discriminate/classify. To this end, AdaptiveMix constructs hard samples by mixing a pair of training images and then narrows down the distance between mixed images and easy training samples represented by the feature extractor for feature space shrinking. 
We evaluate the effectiveness of our AdaptivelyMix with state-of-the-art GAN architectures, including DCGAN \cite{radford2015unsupervised} and StyleGAN-V2 \cite{karras2020analyzing}, which demonstrates that the proposed AdaptivelyMix facilitates the training of GANs and effectively improves the image quality of generated samples.

Besides image generation, our AdaptiveMix can be applied to image classification \cite{zagoruyko2016wide,he2016identity} and Out-Of-Distribution (OOD) detection \cite{hendrycks2018deep,huang2021mos,yu2019unsupervised} tasks, by equipping it with suitable classifiers. To show the way of applying AdaptiveMix, we integrate it with the Orthogonal classifier in recent start-of-the-art work \cite{zaeemzadeh2021out} in OOD. Extensive experimental results show that our AdaptiveMix is simple yet effective, which consistently boosts the performance of \cite{zaeemzadeh2021out} on both robust image classification and Out-Of-Distribution tasks on multiple datasets.

In a nutshell, the contribution of this paper can be summarized as: 
\begin{itemize}
    \item We propose a novel module, namely AdaptiveMix, to improve the training of GANs. Our AdaptiveMix is simple yet effective and plug-and-play, which is helpful for GANs to generate high-quality images.  
    \item  We show that GANs can be stably and efficiently trained by shrinking regions of training data in image representation supported by the discriminator.   
    \item  We show our AdaptiveMix can be applied to not only image generation, but also OOD and robust image classification tasks. Extensive experiments show that our AdaptiveMix consistently boosts the performance of baselines for four different tasks (\emph{e.g.,} OOD) on seven widely-used datasets.
\end{itemize}

\section{Related Work}

\noindent\textbf{The Zoo of Interpolation.}
Since regularization-based methods can simultaneously improve the generalization and robustness with neglectable extra computation costs, this field attracts increasing attention from the community \cite{zhang2017mixup, guo2019mixup, yun2019cutmix}. To some extent, Mixup \cite{zhang2017mixup} is the first study that introduces a sample interpolation strategy for the regularization of Convolution-Neural-Network(CNN)-based models. The virtual training sample, which is generated via linear interpolation with pair-wise samples, smooths the network prediction. Following this direction, many variants were proposed by changing the form of interpolation \cite{verma2019manifold,guo2019mixup,yun2019cutmix,kim2020puzzle,liu2023decoupled}. In this paper, we revisit the interpolation-based strategy and regard the mixed sample as a generated hard sample for shrinking features, which gives a new perspective for the application of mixing operations.  

\noindent\textbf{Regularization in GANs.}
Recently, various methods \cite{salimans2016improved,miyato2018spectral, arjovsky2017wasserstein,jiang2021deceive} attempt to provide regularization on the discriminator to stabilize the training of the GANs. Previous studies designed several penalties, such as weight clipping \cite{arjovsky2017wasserstein} and gradient penalty \cite{gulrajani2017improved} for the parameter of the discriminator. As this might degrade the capacity of the discriminator, spectral normalization is proposed to further stabilize the training via weight normalization. However, spectral normalization has to introduce extra structure, which might limit its application towards arbitrary network architectures. As a more flexible line, adversarial training \cite{zhong2020improving}, C-Reg \cite{zhang2019consistency}, LC-Reg \cite{tseng2021regularizing}, Transform-Reg \cite{mustafa2020transformation}, ADA \cite{karras2020training}, and APA \cite{jiang2021deceive} are proposed in recent years, working similarly to the data augmentation. 
These methods suffer from the sensitivity to the severity of augmented data and have to use adaptive hyper-parameter \cite{karras2020training,jiang2021deceive}. In this paper, we try to address this problem from a new line, that is, training a robust discriminator by shrinking the feature space. Without sacrificing the capability of representation, the proposed method can be elaborated into arbitrary networks and easily combined with existing regularization. Moreover, the proposed method can continually construct hard samples for training without too many hyperparameters, and thus can be used in various additional tasks, such as OOD 
detection \cite{lee2018simple,liang2017enhancing,liu2019large,zaeemzadeh2021out,yang2021semantically} and image classification \cite{krizhevsky2009learning,deng2009imagenet}. 

\section{Method}
In this paper, we investigate how to improve the training of GANs. We first propose a  novel module named AdaptiveMix to shrink the regions of training data in the image representation space of the discriminator. Then, we show that our AdaptiveMix can encourage Lipschitz continuity, and thereby facilitate the performance of GANs. Finally, we equip our AdaptiveMix with an orthogonal classifier of the start-of-the-art OOD method in \cite{zaeemzadeh2021out} to show how to use our module for OOD detection and image recognition tasks. 

\subsection{AdaptiveMix}
\label{sec:DFA}

Our goal is to improve the training of GANs by controlling the discriminator, which can be formulated from the perspective of robust image classification.
Without loss of generality, let the discriminator consist of a \textit{feature extractor} $\mathcal{F}(\cdot)$ and a classifier head $\mathcal{J}(\cdot)$, where the feature extractor is to extract feature from an image, and the classifier is to classify the extracted feature. Our insight is that we can improve the training of GANs by improving the representation of the feature extractor $\mathcal{F}$, motivated by studies on robust image representation \cite{verma2019manifold,Zhuge_2021_CVPR}. 
As observed in \cite{verma2019manifold}, vanilla classification networks scatter  training data in their feature  space, driving  the classifier  improperly to assign high confidences to samples that are off the underlying manifold of training data. Similarly, with such representation,  it is difficult for the discriminator  to learn the distribution of real data. 
Therefore, we propose to shrink the regions of training data in the
image representation space supported by the feature extractor of the discriminator for improving the training of GANs.

We propose a module, termed AdaptiveMix, to shrink the regions of training data in the space represented by a feature extractor $\mathcal{F}$. However, it is intractable to directly capture the regions of training data in the   feature space.  Given training samples of a class $c$,  our insight is that we can shrink its regions in the feature space by reducing the distance between hard and easy samples in the feature space, where hard samples are regarded as samples that are difficult for networks to classify. In other words, we argue that most hard samples are more peripheral than easy ones in the feature space formed by all training samples of a class, which leads the decision boundaries to enlarge the intra-class distance for covering the hard samples. Therefore,  for class $c$,  if we pull hard samples towards easy samples, the regions of  training samples of class $c$ can be shrunk in the feature space. 
Therefore, the proposed AdaptiveMix consists of two steps. First, AdaptiveMix generates hard samples from the training data. Second, our AdaptiveMix reduces  the distance between hard  and easy samples. 

\begin{figure}[tbp]
    \centering
    \includegraphics[width=.46\textwidth]{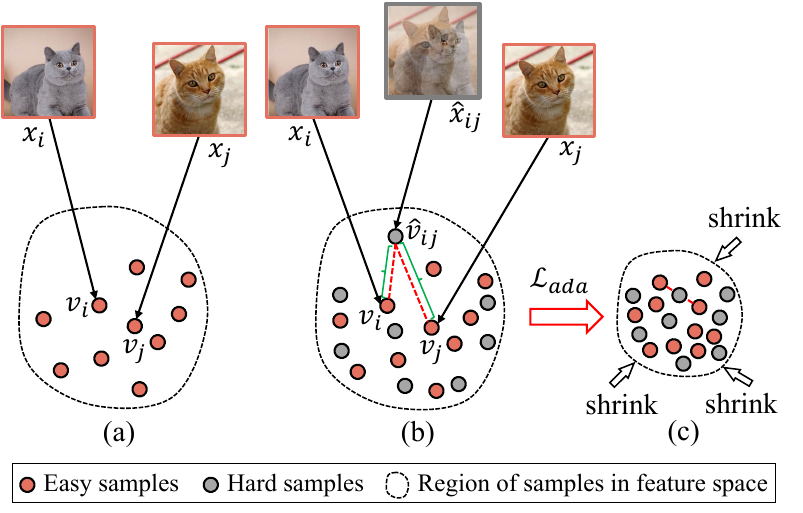}
    \caption{The illustration of our AdaptiveMix. (a) Easy samples $x_i$  and $x_j$ of a class are projected into the  feature space by feature extractor $\mathcal{F}(\cdot)$, where $v_i=\mathcal{F}(x_i)$.  
    (b) Hard  sample $\hat{x}_{ij}$ is generated by the linear combination of a training sample pairs $x_i$  and $x_j$,  and is projected to the feature space. 
    (c) AdaptiveMix shrinks the region of the class  in the feature space by reducing the feature distance between easy and hard samples. 
    }

    \label{fig:hard_sample}
    \vspace{-8pt}
\end{figure}

\noindent \textbf{Hard Sample Generation.} A naive manner of finding hard samples is to employ  trained networks to evaluate training samples, where samples to which the networks assign the prediction with low confidence are considered as hard ones.  However, this introduces new issues. For example,  this requires  well-trained networks,  which are not always available. Instead, we propose a simple way to generate hard samples, inspired by the promising performance of Mixup-based image augmentation methods \cite{zhang2017mixup,guo2019mixup,hong2021stylemix}.  Recently, various Mixup-based methods were proposed to mix multiple images into a new image. Here we employ the vanilla version of Mixup \cite{zhang2017mixup} to generate hard samples for simplicity. 
Let $\mathcal{X}=\{x_i\}_{i=1}^N$ denote $N$ training images, where $x_i$ is the $i$-th training image.  AdaptiveMix mixes a pair of training images  to generate a hard sample following Mixup: 
\vspace{-10pt}
\begin{align}
    \hat{x}_{ij} = g(x_i,x_j, \lambda) = \lambda x_i + (1 - \lambda) x_j,
\end{align}
where $g(\cdot,\cdot, \lambda)$ is a function  linearly  combining $x_i$ and $x_j$,  $\lambda$ is a hyper-parameter sampled from Beta distribution $\lambda \in \mathbb{B}(\alpha,\alpha)$. 

As shown in Fig. \ref{fig:hard_sample},  mixed sample $\hat{x}_{ij}$  is more confusing and difficult for networks to discriminate, compared with original training samples $x_i$ and $x_j$. Without loss of generalization, we refer to  an original training sample $x_i$ as an easy sample, and $\hat{x}_{ij} $   as a hard one.

\noindent \textbf{AdaptiveMix Loss}. Our AdaptiveMix reduces the distance between a hard sample $\hat{x}_{ij}$  and its corresponding  easy ones $x_i$ and $x_j$  in the representation space represented by feature extractor $\mathcal{F(\cdot)}$, in order  to shrink regions of training data of classes that $x_i$ or $x_j$ belongs to. Note that we propose a soft loss, since   hard sample $\hat{x}_{ij}$ does not completely belong to the class of $x_i$ or $x_j$. We reduce the distance between  $\hat{x}_{ij}$ and $x_i$ in the feature space according to the proportion of $x_i$ in   $\hat{x}_{ij}$ in the linear combination: 
\begin{align}
\label{eq:la}
  \mathcal{L}_{ada} \!\!=\!\! \sum_i\sum_j \mathbb{D}_v(\lambda \mathcal{F}(x_i) \!+\! (1-\lambda) \mathcal{F}(x_j),  \mathcal{F}(\hat{x}_{ij}) \!+\! \sigma),
\end{align}
where $\sigma$ is a noise term sampled from Gaussian distribution to prevent over-fitting and $\mathbb{D}_v(\cdot,\cdot)$ refers to the metric to evaluate the distance, like L1 norm, L2 norm. 
Note that our AdaptiveMix does not need labels of training images; however, it is able to shrink the regions of training data  for  each class in the feature space (see Fig. \ref{fig:hard_sample}), since easy sample $x_i$ and its associated hard one  $\hat{x}_{ij}$ belong to the same class. 

\subsection{Connections to Lipschitz Continuity}
To further investigate the superiority of our method, we theoretically analyze the relationship between the proposed AdaptiveMix and Lipschitz continuity. 

\vspace{1mm}
\noindent \textbf{Preliminary.} In the proposed method, the feature extractor $\mathcal{F}(\cdot)$ connects the input space $\mathcal{X}$ and the embedding space $\mathcal{V}$. Given two evaluation metrics $\mathbb{D}_x(\cdot,\cdot)$ and $\mathbb{D}_v(\cdot,\cdot)$ defined on $\mathcal{X}$ and $\mathcal{V}$, respectively, $\mathcal{F}(\cdot)$ fulfills Lipschitz continuity if a real constant $K$ exists to ensure all $x_i,x_j \in \mathcal{X}$ meet the following condition: 
\begin{align}
\label{eq:lip}
    K \mathbb{D}_x(x_i,x_j) \geq  \mathbb{D}_v(\mathcal{F}(x_i),\mathcal{F}(x_j)).
\end{align}

\vspace{1mm}
\noindent \textbf{Proposition.} Based on the analysis in \cite{verma2019manifold,cisse2017parseval}, a flat embedding space, especially with Lipschitz continuity, is an ideal solution against unstable training and adversarial attack. Hence, the effectiveness of the proposed method can be justified by proving the equivalence between AdaptiveMix and $K$-Lipschitz continuity.

\vspace{3mm}
\noindent \textbf{Theorem.} Towards any $K$ of Lipschitz continuity, AdaptiveMix is an approximate solution under L1 norm metric space.

\vspace{3mm}
\noindent \textbf{Proof.} Given $x_i$ and $x_j$ sampled from $\mathcal{X}$, their linear combination based pivot $\hat{x}_{ij}$ can be obtained via $g(x_i,x_j,\lambda)$. Since $\hat{x}_{ij}$ can be regarded as a sample in $\mathcal{X}$, we can transform Lipschitz continuity (Eq. (\ref{eq:lip})) to
\begin{equation}
\begin{aligned}
        K\mathcal{B} =  &  K (\mathbb{D}_x(\hat{x}_{ij}, \lambda x_i) + \mathbb{D}_x(\hat{x}_{ij}, (1-\lambda) x_j))\\
          & \geq  \mathbb{D}_v(\hat{v}_{ij}, \lambda v_i) + \mathbb{D}_v(\hat{v}_{ij}, (1-\lambda) \lambda v_j) \\
          & \geq \mathbb{D}_v(\hat{v}_{ij}, g(v_i,v_j,\lambda))
\end{aligned}
\end{equation}
where $\mathbb{E}[\lambda]=0.5$ and $\mathbb{E}[x_i] = \mathbb{E}[x_j]$. Hence the upper bound $\mathcal{B}$ can be estimated as 0 through mini-batch training. As $\mathbb{D}_v(\cdot,\cdot)$ is an L1-norm distance, $\mathbb{D}_v(\cdot,\cdot)$ should be no less than 0. Hence, we can get the lower and upper bound of $\mathbb{D}_v(\hat{v}_{ij}, g(v_i,v_j,\lambda))$ within Lipschitz continuity:
\begin{align}
    \mathbb{E}[0] \leq \mathbb{E}[\mathbb{D}_v(\hat{v}_{ij},g(v_i,v_j,\lambda)))] \leq \mathbb{E}[K\mathcal{B}] = 0.
\end{align}
Therefore, if $\mathcal{F}$ is under the Lipschitz continuity, $\mathbb{D}_v(\hat{v}_{ij},g(v_i,v_j,\lambda)))$ should be zero, and the optimal result of AdaptiveMix is identical to $\mathbb{D}_v(\hat{v}_{ij}, \Lambda_{\lambda}(v_i,v_j))$. Therefore, $K$-Lipschitz continuity can be ensured by minimizing AdaptiveMix.

\noindent \textbf{Intuition.}
This theoretical result is also consistent with intuition. Our general idea is to shrink the feature space for robust representation. As Lipschitz continuity requires that the distance in embedding space should be lower than that in image space, shrinking feature space should be a reasonable way to approximately ensure Lipschitz continuity. 

\subsection{AdaptiveMix-based Image Generation}
Based on the previous analysis, the proposed module  can help to stabilize the training of GANs.
In this paper, we show how to apply AdapativeMix to image generation by integrating it with two state-of-the-art image generation methods,  WGAN \cite{arjovsky2017wasserstein}, and StyleGAN-V2 \cite{karras2020analyzing}.  We mainly elaborate on the integration of  WGAN  in the main paper, and that of StyleGAN-V2 is given in the \textit{supplementary materials}. 
Thanks to the  plug-and-play property of   AdaptiveMix, we  equip WGAN  with AdaptiveMix  in a simple manner.  In particular,   we apply  AdaptiveMix to WGAN's discriminator  consisting of feature extractor $\mathcal{F}(\cdot)$ and classifier head $\mathcal{J}(\cdot)$
 We then rewrite the learning objective to  add the  AdaptiveMix  to WGANs:

\begin{equation}
\begin{aligned}
    \min_{G} \max_{\mathcal{F},\mathcal{J}} \!\!\underset{x \sim p_r}{\mathbb{E}}\!\!\![\mathcal{J}(\mathcal{F}(x))]\!\!-\!\!\!\!\underset{z \sim p_z}{\mathbb{E}}[\mathcal{J}(\mathcal{F}(G(z)))] \\
    + \min_{G,\mathcal{F}}\underset{x \sim p_r,p_g}{\mathbb{E}} [\mathcal{L}_{ada}]
\end{aligned}
\end{equation}
where $z$ is the noise input of the generator $G(\cdot)$; L2 norm is adapted as the metrics for $\mathcal{L}_{ada}$,  the output of $\mathcal{J}(\cdot)$ refers to a scalar to estimate the realness of the given sample. 
 To simplify the structure of $\mathcal{J}(\cdot)$, we directly adopt averaging operator as $\mathcal{J}(\cdot)$.  In this paper, AdaptiveMix generates hard samples by the linear combination of  real samples and fake ones generated by the generator. Such mixing is a kind of feature smoothing that enforces the decision boundaries of the  discriminator to be smooth, improving the training stability of GANs.
The pseudo-code is given in \textit{supplementary material}. Note that the traditional mixing-based methods do \textbf{not} works for the zoo of WGANs, since  WGAN plays a dynamic min-max game where the output of the discriminator ranges from ($-\infty,+\infty$), while our method improves the training of WGAN.

\subsection{AdaptiveMix-based Image Classification}

Besides image generation, the proposed AdaptiveMix can be applied to image classification. Here we show how to apply AdativeMix to this task. Different from the image generation task,  image classification  requires  features extracted by  the feature extractor $\mathcal{F}(\cdot)$  of a classification model to be discriminative as much as possible.  Our AdaptiveMix shrinks regions of training data in the  feature space, which smooths  features to some extent if AdaptiveMix is solely applied. Nevertheless, this can be easily addressed  by adopting a proper classifier that enforces features of different classes to be separable.

Inspired by image classification method \cite{zaeemzadeh2021out}, we employ an orthogonal classifier $\tilde{\mathcal{J}}(\cdot)$ to ensure the class-aware separation in the feature space, where the orthogonal classifier $\tilde{\mathcal{J}}(\cdot)$ consists of several weight vectors $w_k \in \mathcal{W}$, and $w_k$ corresponding to the $k$-th class. 
In particular, we replace the last fully-connected layer of a CNN-based classification model with the orthogonal classifier $\tilde{\mathcal{J}}(\cdot)$. 
Thus, given  $x_i$, the  prediction score $y^k$  to $k$-th class is calculated as:
\begin{align}
    y^k_i = \frac{w_k^T v_i}{||w_k||\; ||v_i||}.
\end{align}
where $v_i = \mathcal{F}(x_i)$ denoted $x_i$'s  feature extracted by feature extractor $\mathcal{F}(\cdot)$ of the classification model.
 The probability $p^k_i$ that $x_i$ belongs to $k$-th class is  calculated via a softmax layer:
\begin{align}
    p^k_i = \frac{exp{(y^k_i})}{\sum\limits_{1 \leq l \leq n} exp({y^l_i})}
\end{align}
where the set $\mathcal{P}_i = \tilde{\mathcal{J}}(v_i) = \{ p^k_i| 1 \leq k \leq n \}$ forms the final output of the CNN-based model for an $n$-class recognition task. 

By removing the bias and activation function in the last layer, the classification model maps $x$ into the allowed norm ball space, which ensures that features corresponding to different classes  can be  separable. To further strengthen the class-aware separation, we then introduce the orthogonal constraint to initialize $\mathcal{W}$, which is defined as:
\begin{align}
    \label{eq:or}
    \prod_{w_k,w_l \in \mathcal{W},k\neq l} w_k^T w_l = 0.
\end{align}
In addition,  besides AdaptiveMix loss,  we can use mixing-based cross-entropy loss in the learning objective of image classification following augmentation \cite{zhang2017mixup}, since we use Mixup to generate hard samples (See \textit{supplementary material} for more details on classification).

\subsection{AdaptiveMix-based OOD Detection}
 AdaptiveMix can be easily integrated into the state-of-the-art OOD detection model of \cite{zaeemzadeh2021out}. Given all training samples $x \in \mathcal{X}$ as input, we can obtain the corresponding representation $v \in \mathcal{V}$ via the trained $\mathcal{F}(\cdot)$. Then, the representative representation $v^*_k$ of the $k$-th class can be obtained by computing the first singular vectors of $\{\mathcal{F}(x_i)| x_i \in \mathcal{X}\; \mathop{\arg\max}\widetilde{y_i} = k \}$. Note that $v^*_k$ is calculated by SVD rather than $\mathcal{F}(x_i)$. Given a test sample $x_t$, the probability  $\phi_t$  that $x_t$ is OOD is  calculated as:
\begin{align}
    \phi_t = \min_k \text{arccos}(\frac{|\mathcal{F}^T(x_t) v_k^*|}{||\mathcal{F}(x_t)||}) ,
\end{align}
where $x_t$ is categorized as an OOD sample if $\phi_t$ is larger than a predefined threshold $\phi^*$.

\section{Experiments}

\begin{table}[]
\caption{Summary of  improvements by using our AdaptiveMix, where   \textbf{Gain} refers to our improvement over the baselines.  Our method AdaptiveMix boosts  the performance of six baselines across four tasks on seven  widely-used datasets.  Detailed comparison results are provided in tables specified in the \em{Table \#} column. }
\label{tab:overview_results}
\resizebox{.48\textwidth}{!}{
\begin{tabular}{c|c|c|
>{\columncolor[HTML]{EFEFEF}}c }
\hline
& Dataset & Table \# & \textbf{Gain}                    \\ \hline
& C-10 \cite{krizhevsky2009learning}               & Table \ref{tab:compared}   & \textbf{-20.0\% FID}  $\downarrow$                    \\ 
& CelebA  \cite{liu2015deep}             & Table \ref{tab:compared}   & \textbf{-54.0\% FID} $\downarrow$                     \\ 
& FFHQ  \cite{karras2019style}             & Table \ref{tab:stylegan}   & \textbf{-4.9\% FID}  $\downarrow$                     \\ 
& AFHQ-CAT \cite{choi2020stargan}          & Table \ref{tab:stylegan}   & \textbf{-43.5\% FID}  $\downarrow$                    \\
\multirow{-5}{*}{\begin{tabular}[c]{@{}c@{}}Image \\ Generation\end{tabular}}      & FFHQ-5k  \cite{karras2019style}             & Table .\ref{tab:ffhq-5k}   & \textbf{-47.9\% FID}   $\downarrow$                    \\ \hline \hline
& C-10  \cite{krizhevsky2009learning}         & Table \ref{tab:cleanacc}   & \textbf{+0.7\% Acc.} $\uparrow$            \\ 
& C-100  \cite{krizhevsky2009learning}         & Table \ref{tab:cleanacc}   & \textbf{+1.5\% Acc.} $\uparrow$              \\ 
& T-ImageNet \cite{deng2009imagenet}        & Table \ref{tab:cleanacc}   & \textbf{+5.87\% Acc.} $\uparrow$  \\
\multirow{-4}{*}{\begin{tabular}[c]{@{}c@{}}Image \\ Classification\end{tabular}}  & ImageNet \cite{russakovsky2015imagenet}        & Table \ref{tab:cleanacc}   & \textbf{+1.9\% Acc.}  $\uparrow$                      \\ \hline 
& C-10 \cite{krizhevsky2009learning}          &Table \ref{tab:ablation study}   & \textbf{+4.6$\times$ Acc.} $\uparrow$  \\ 
& C-100 \cite{krizhevsky2009learning}          & Table \ref{tab:adv_cifar100_tiny}   & \textbf{+5.2$\times$ Acc.} $\uparrow$   \\ 
\multirow{-3}{*}{\begin{tabular}[c]{@{}c@{}}Robust \\ Classification\end{tabular}} & T-ImageNet \cite{deng2009imagenet}        & Table \ref{tab:adv_cifar100_tiny}   & \textbf{+1.1 $\times$ Acc.} $\uparrow$                \\ \hline \hline
\begin{tabular}[c]{@{}c@{}}OOD\\ Detection\end{tabular}                            & Benchmark \cite{zaeemzadeh2021out}         & Table \ref{tab:OOD}   & \textbf{+3.5\% F1}  $\uparrow$                \\ \hline
\end{tabular}
}
\vspace{-2pt}
\end{table}

To evaluate the performance of our method, we conduct extensive experiments on various tasks, including image generation, image recognition, robust image classification, and OOD detection.  Table \ref{tab:overview_results} shows that the proposed method improves the baselines significantly on these tasks. Below, we first evaluate the performance of the proposed method on image generation and then test the proposed method on visual recognition tasks such as image classification and OOD detection. Note that we elaborate on datasets used in the experiments, additional experiments, and  implementation details in the \textit{supplementary material}.  

\subsection{Performance on Image Generation}
\noindent {\bf Ablation Study.}
To quantify the contribution of AdaptiveMix, we test the image generation performance with or without  AdaptiveMix. As listed in Table \ref{tab:ablation_img}, the proposed method outperforms the baseline \cite{arjovsky2017wasserstein} significantly in all cases. Such improvement indicates that the proposed method  boosts the training of GANs.

In addition to the quantitative analysis, we also conduct visualizations to showcase the effectiveness of the proposed method. As shown in Fig. \ref{fig:toys}, a toy dataset is used to train the model, where the input is the extensive 2D points following the distribution in Fig. \ref{fig:toys}. As can be seen, AdaptiveMix can mimic such distribution better than the other two baselines. For further investigation, the output of the discriminator for each position is visualized in the second row of Fig. \ref{fig:toys}. If the discriminator enlarges the distance between real and generated samples like Std-GAN \cite{goodfellow2014generative}, the generator is hard to derive useful guidance from the discriminator, leading to poor generated results or mode collapse. In contrast, the proposed method shrinks the input samples, hence the confidence score map is flattened, which can provide more health gradient for the generator, resulting in a better generation performance. We also quantify such a phenomenon in the practical case. As shown in Table \ref{tab:lipschitz}, we calculate the averaging ratios between the distances in the feature space and image space.  By adopting AdaptiveMix, the corresponding ratios are minimized among all the cases, which can be regarded as a guarantee for the Lipschitz continuity.

\begin{table}[!tb]
\caption{The ablation study of the proposed method on CIFAR10 \cite{krizhevsky2009learning}. Mean FID $\pm$ S.D. refers to the mean and standard deviation of FID scores when the models are trained by 100, 400, and 800 epochs. Min FID is the optimal result during training. The convergence curve is given in the \textit{supplementary materials} }
\label{tab:ablation_img}
\centering
\vspace{-5pt}
\setlength\tabcolsep{16pt}
\resizebox{.48\textwidth}{!}{
\begin{tabular}{c|c|c}
\hline
                 & \textbf{Mean FID $\pm$ S.D.} $\downarrow$ & \textbf{Min FID} $\downarrow$  \\ \hline
Baseline             &   106.81 $\pm$ 21.79 &   55.96   \\ \hline \hline
\rowcolor[HTML]{EFEFEF}
Baseline + AdaptiveMix           &\textbf{39.52 $\pm$ 7.85}&  \textbf{30.85} \\ \hline
\end{tabular}
}
\end{table}

\begin{figure*}[]
    \centering
    \includegraphics[width=0.95\textwidth]{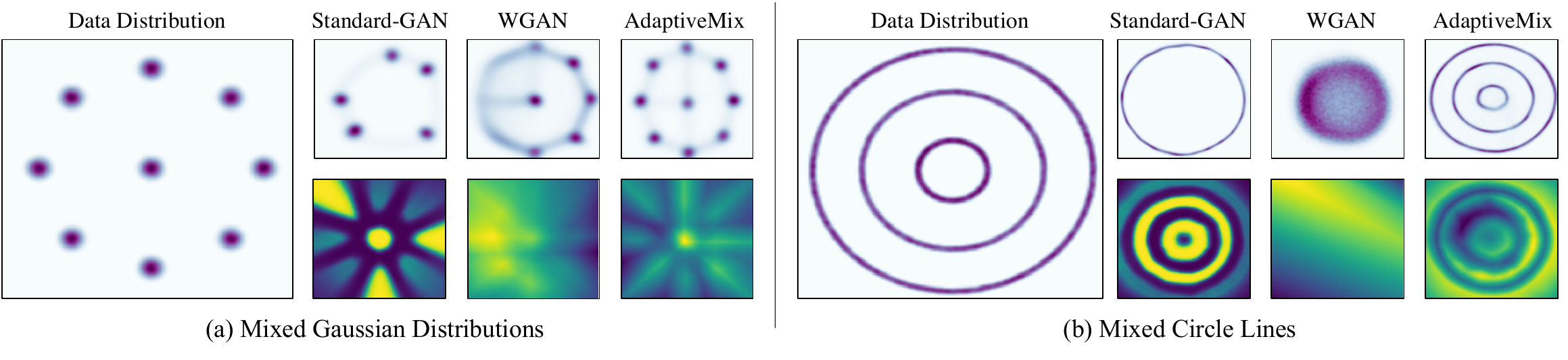}
    \caption{The experimental results on a synthetic data set:  2D points from (a) nine Gaussian distributions and (b) three circles are adopted as the training data for GANs. In each case: From left to right: the samples generated by Std-GAN, WGAN and AdaptiveMix. The first row refers to the generated results and the second row is the corresponding confidence map of the discriminator.}
    \label{fig:toys}
    \vspace{-9pt}
\end{figure*}

\begin{table}[]
\caption{ The evaluation for the Lipschitz continuity on FFHQ-5k \cite{karras2019style} with StyleGAN-V2 \cite{karras2020analyzing}. The criterion refers to the averaging of $\frac{\text{embedding distance}}{ \text{image distance}}$ on the given pairs of samples, \emph{i.e.} $\frac{\mathbb{D}_v(\mathcal{F}(x_i),\mathcal{F}(x_j))}{\mathbb{D}_x(x_i,x_j)}$. Smaller is better. The details of the calculation can be found in \textit{supplementary material.}} 
\label{tab:lipschitz}
\vspace{-5pt}
\resizebox{.48\textwidth}{!}{
\begin{tabular}{c|c|c|c}
\hline
                   & Real Samples    & Generated Samples & Both           \\ \hline
StyleGAN-V2        & 3.167          & 0.263            & 0.735          \\ \hline
\rowcolor[HTML]{EFEFEF} 
StyleGAN-V2 + Ours & \textbf{1.391} & \textbf{0.166}   & \textbf{0.291} \\ \hline
\end{tabular}
}
\vspace{-10pt}
\end{table}

\begin{table}[]
\centering
\caption{ FIDs of  DCGAN \cite{radford2015unsupervised} using  various learning objectives  on CelebA \cite{liu2015faceattributes} and CIFAR10 \cite{krizhevsky2009learning}. }
\label{tab:compared}
\vspace{-5pt}
\resizebox{.48\textwidth}{!}{
\setlength\tabcolsep{4pt}
\begin{tabular}{c|cc}
\hline
Learning Objective             & CIFAR-10                       & CelebA                 \\ \hline
WGAN \cite{arjovsky2017wasserstein}  (\small ICML'17)                        & 55.96                          &   -                    \\
HingeGAN \cite{zhao2016energy}  (\small ICLR'17)                      & 42.40                          & 25.57                  \\
LSGAN \cite{mao2017least} (\small ICCV'17)                        & 42.01                           & 30.76                  \\
DCGAN \cite{radford2015unsupervised}   (\small ICLR'16)                       & 38.56                          & 27.02                  \\ \hline\hline
WGAN-GP \cite{gulrajani2017improved}        (\small NIPS'17)               & 41.86                          & 70.28                  \\
Re-implemented WGAN-GP  & 38.63                          & 70.16                  \\ \hline\hline
Realness GAN-Obj.1 \cite{xiangli2020real} (\small ICLR'2020)            & 36.73                          & -                      \\
Realness GAN-Obj.2  \cite{xiangli2020real} (\small ICLR'2020)            & 34.59                          & \underline{23.51}                  \\
Realness GAN-Obj.3  \cite{xiangli2020real}   (\small ICLR'2020)          & 36.21                          & -                      \\ \hline\hline
\rowcolor[HTML]{EFEFEF}
AdaptiveMix (Ours)                & \textbf{30.85}                 &     \textbf{12.43}                   \\ \hline
\end{tabular}
}
\vspace{-3pt}
\end{table}

\noindent {\bf Comparing with Existing Methods.}
To further show the superiority of the proposed method, we compare the performance of the proposed method with the well-known  loss functions  on the toys dataset, CIFAR10 \cite{krizhevsky2009learning}, and CelebA \cite{liu2015faceattributes}. As shown in Table \ref{tab:compared}, the proposed method outperforms other methods in both datasets by a large margin. Compared with the recent method, Realness GAN, a 10.81\% improvement in FID is achieved by the proposed method on CIFAR10. Similarly, in the case of CelebA, the FIDs of Realness GAN and the proposed method are 23.51 and 12.43, respectively, which convincingly shows the advantage of AdaptiveMix. Note that the results in Table \ref{tab:compared}  are taken from \cite{xiangli2020real} based on the same architecture, i.e. DCGAN. The corresponding visualization results are given in the \textit{supplementary material}.

\begin{table}[]
\centering
\caption{ FID and IS of the proposed method on AFHQ \cite{choi2020stargan} and FFHQ \cite{karras2019style} for StyleGAN-V2 \cite{karras2020analyzing} compared with the other state-of-the-art solutions for GAN training}
\label{tab:stylegan}
\vspace{-5pt}
\resizebox{.48\textwidth}{!}{
\begin{tabular}{c|cc|cc}
\hline
                    & \multicolumn{2}{c|}{AFHQ-Cat-5k}  & \multicolumn{2}{c}{FFHQ (Full)} \\ \cline{2-5} 
\multirow{-2}{*}{Method}  & FID             & IS                        & FID             & IS             \\ \hline
StyleGAN-V2 \cite{karras2020analyzing}  ({\small CVPR'20})       & 7.737           & 1.825                 & 3.862           & 5.243          \\ 
StyleGAN-V2 (Re-Impl.)       & 7.924           & 1.890                 & 3.810           & 5.185          \\ \hline
LC-Reg \cite{tseng2021regularizing} ({\small CVPR'21})            & 6.699           & 1.943             & 3.933           & \textbf{5.312}          \\ \hline 
\rowcolor[HTML]{EFEFEF} 
Style GAN-V2 + Ours & \textbf{4.477}           & \textbf{1.972}                       & \textbf{3.623}                & 5.222               \\ \hline \hline
ADA \cite{karras2020training}   ({\small NIPS'20})              & 6.053           & 2.119         & 4.018           & 5.329          \\
ADA  (Re-Impl.)              & 5.582           & 2.059         & 3.713          &5.200           \\
\rowcolor[HTML]{EFEFEF} 
ADA + Ours          & \textbf{4.680}            & \textbf{2.069}                          & \textbf{3.681}                & \textbf{5.335}               \\ \hline \hline
APA  \cite{jiang2021deceive} ({\small NIPS'2021})               & 4.876           & 2.156              & 3.678           & 5.336          \\
APA (Re-Impl.)               & 4.645           & 2.093             & 3.752           & 5.281          \\
\rowcolor[HTML]{EFEFEF} 
APA+Ours            & \textbf{4.148}           & \textbf{2.096}                       &  \textbf{3.609}               &  \textbf{5.296}             \\ \hline
\end{tabular}
}
\vspace{-3pt}
\end{table}
\begin{table}[]
\caption{FID and IS of our method compared to previous techniques for regularizing GANs on FFHQ-5k \cite{karras2019style}. StyleGAN-V2 \cite{karras2020analyzing} is used as the baseline.}
\label{tab:ffhq-5k}
\vspace{-5pt}
\setlength\tabcolsep{16pt}
\resizebox{.48\textwidth}{!}{
\begin{tabular}{c|c|c}
\hline
Regularization    & FID             & IS             \\ \hline
Baseline  \cite{karras2020analyzing}  ({\small CVPR'20})        & 37.830          & 4.018          \\
Baseline  (Re-Impl.)        & 36.053          & 4.097          \\ \hline
Instance Noise \cite{sonderby2016amortised} ({\small ICLR'17})   & 40.981          & 4.231          \\
One-sided LS \cite{salimans2016improved} ({\small NIPS'16}  )      & 33.978          & 4.029          \\
LC-Reg    \cite{tseng2021regularizing} ({\small CVPR'21})         & 35.148          & 3.926          \\ \hline
\rowcolor[HTML]{EFEFEF} 
Ours     & \textbf{18.769} & \textbf{4.332} \\ \hline \hline
APA  \cite{jiang2021deceive} ({\small NIPS'2021})              & 13.249          & 4.487          \\ 
APA (Re-Impl.)              & 14.368          & 4.855          \\ \hline
\rowcolor[HTML]{EFEFEF} 
APA+Ours & \textbf{11.498} & \textbf{4.866} \\ \hline
\end{tabular}
}
\vspace{-10pt}
\end{table}

In order to comprehensively justify AdaptiveMix, we also compare the proposed method with the recent regularization for GANs. Table \ref{tab:stylegan} shows the proposed method can help the convergence of GANs on different datasets  and achieve remarkable results. As a plug-and-play module, the proposed method can also be combined with the state-of-the-art augmentation-based methods, ADA \cite{karras2020training} and APA \cite{jiang2021deceive}, which further improves the generation performance of GANs ( 13.5\% improvement in FID averagely). Finally, we evaluate the performance of the proposed method using limited training data. As shown in Table \ref{tab:ffhq-5k}, given only 5k samples, the proposed method can significantly improve the baseline from 37.830 to 18.769 in FID. By combining with APA, AdaptiveMix can achieve the best FID (11.498) and IS (4.866) scores.

\subsection{Performance on Visual Recognition}
\label{sec:adversarial}

\begin{figure}[!htb]
    \centering
    \includegraphics[width=.45\textwidth]{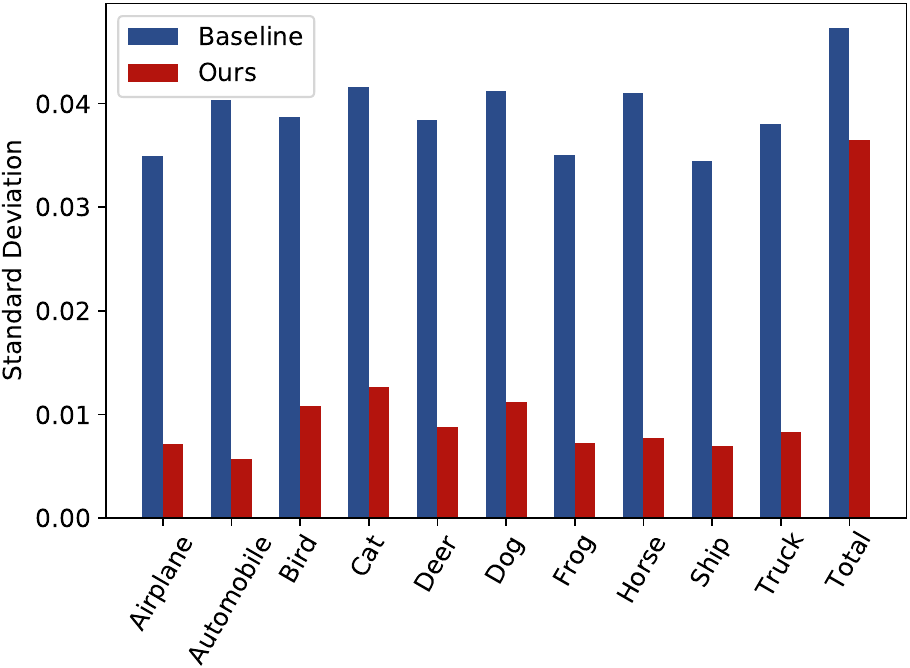}
    \caption{Compactness (\emph{i.e.,} standard deviation) of the embedding clusters on CIFAR-10. Standard deviation is calculated on the embedding codes within the same annotation. The `Total' is the compactness of the whole embedding codes in the test set.}
    \label{fig:results}
    \vspace{-5pt}
\end{figure}

\begin{table}[!htbp]
\centering
\caption{Accuracy (\%) on CIFAR-10 based on WRN-28-10 trained with the various methods with orthogonal classifier (Orth.).}
\label{tab:ablation study}
\Huge
\vspace{-5pt}
\resizebox{.48\textwidth}{!}{
\begin{tabular}{c|ccccc}
\hline
CIFAR10                               & \begin{tabular}[c]{@{}c@{}}FGSM \\ (8/255)\end{tabular} & \begin{tabular}[c]{@{}c@{}}PGD-8\\ (4/255)\end{tabular} & \begin{tabular}[c]{@{}c@{}}PGD-16\\ (4/255)\end{tabular} & \begin{tabular}[c]{@{}c@{}}CW-100\\ (c=0.01)\end{tabular} & \begin{tabular}[c]{@{}c@{}}CW-100\\ (c=0.05)\end{tabular}             \\ \hline
Baseline                              & 38.03                                                   & 0.92                                                    & 0.28                                                     & 11.1                                                      & 0.39                                                               \\\hline
Mixup \cite{zhang2017mixup} & 60.17 & 3.97 & 1.16 & 30.32 & 2.36 \\
Orth. + Mixup                    & 44.80                                                   & 3.99                                                    & 2.66                                                     & 71.12                                                     & 49.47                                                              \\ \hline
M.-Mixup \cite{verma2019manifold} & 59.32 & 7.97 & 2.97 & 51.47 & 11.12 \\
Orth. + M.-Mixup           & 38.76                                                   & 5.77                                                    & 4.38                                                     & 69.08                                                     & 53.98                                                               \\ \hline \hline
\rowcolor[HTML]{EFEFEF} 
Ours                                  & \textbf{74.18}                                          & \textbf{32.12}                                          & \textbf{22.12}                                           & \textbf{81.39}                                            & \textbf{74.72}                                             \\ \hline
\end{tabular}
}
\vspace{-4pt}
\end{table}

\noindent {\bf Ablation Study.} 
To validate that our method shrinks the regions of samples in the feature space, we also analyze the cluster compactness (\emph{i.e.,} standard deviation of the cluster) of each class in the feature space on CIFAR-10, which is presented in Fig. \ref{fig:results}. It can be observed that the class-wise standard deviation of our AdaptiveMix is much lower than that of the baseline. The entry of `Total' measures the compactness of regions of all samples in the feature space. Fig. \ref{fig:results} shows that our method shrinks regions of samples in the feature space,  compared with that of the baseline (\emph{i.e.,} without AdaptiveMix).

\begin{table}[]
\caption{Accuracy (\%) on CIFAR-100 and Tiny-ImageNet against various adversarial attacks based on WRN-28-10\cite{zagoruyko2016wide} and PreActResNet-18\cite{he2016identity} respectively.}
\label{tab:adv_cifar100_tiny}
\Huge
\vspace{-5pt}
\resizebox{.48\textwidth}{!}{
\begin{tabular}{c|c|ccccc}
\hline
Dataset & Method   & \begin{tabular}[c]{@{}c@{}}FGSM\\ (8/255)\end{tabular} & \begin{tabular}[c]{@{}c@{}}PGD-8\\ (4/255)\end{tabular} & \begin{tabular}[c]{@{}c@{}}PGD-16\\ (4/255)\end{tabular} & \begin{tabular}[c]{@{}c@{}}CW-100\\ (c=0.01)\end{tabular} & \begin{tabular}[c]{@{}c@{}}CW-100\\ (c=0.05)\end{tabular} \\ \hline
\multirow{4}*{C-100} &Baseline  & 11.71                                                  & 0.79                                                    & 0.42                                                     & 4.42                                                      & 0.23                                                      \\ \cline{2-7}
&Mixup \cite{zhang2017mixup}    & 27.34                                                  & 0.28                                                    & 0.11                                                     & 4.83                                                      & 0.28                                                      \\ \cline{2-7}
&M.-Mixup \cite{verma2019manifold} & \textbf{29.73}                                         & 1.19                                                    & 0.49                                                     & 10.75                                                     & 0.77                                                      \\ \cline{2-7}

&\cellcolor[HTML]{EFEFEF}Ours     & \cellcolor[HTML]{EFEFEF}24.28                                                  & \cellcolor[HTML]{EFEFEF}\textbf{8.22}                                           & \cellcolor[HTML]{EFEFEF}\textbf{7.40}                                            & \cellcolor[HTML]{EFEFEF}\textbf{42.02}                                            & \cellcolor[HTML]{EFEFEF}\textbf{26.18}                                            \\ \hline\hline
\multirow{4}*{T-ImageNet }&Baseline  & 4.26                                                   & 0.81                                                    & 0.60                                                     & 27.92                                                     & 7.52                                                      \\ \cline{2-7}
&Mixup \cite{zhang2017mixup}   & 4.23                                                   & 0.98                                                    & 0.77                                                     & 29.13                                                     & 15.41                                                     \\ \cline{2-7}
&M.-Mixup \cite{verma2019manifold} & 3.04                                          & 0.82                                                    & 0.59                                                     & 29.69                                                     & 16.86                                                     \\ \cline{2-7}

&\cellcolor[HTML]{EFEFEF} Ours     & \cellcolor[HTML]{EFEFEF} \textbf{7.10}                                          & \cellcolor[HTML]{EFEFEF} \textbf{4.66}                                           & \cellcolor[HTML]{EFEFEF} \textbf{4.98}                                            & \cellcolor[HTML]{EFEFEF} \textbf{35.93}                                            & \cellcolor[HTML]{EFEFEF} \textbf{34.22}                                            \\ \hline

\end{tabular}
}
\vspace{-10pt}
\end{table}

\begin{table}[!h]
\vspace{-4pt}
\centering
\caption{Accuracy (\%) on CIFAR-10 based on WRN-28-10 trained with the proposed method under various noise terms $\sigma$.}
\label{tab:noise}
\vspace{-5pt}
\setlength\tabcolsep{6pt}
\resizebox{.48\textwidth}{!}{
\begin{tabular}{c|ccccc}
\hline
Noise  & \begin{tabular}[c]{@{}c@{}}FGSM \\ (8/255)\end{tabular} & \begin{tabular}[c]{@{}c@{}}PGD-8\\ (4/255)\end{tabular} & \begin{tabular}[c]{@{}c@{}}PGD-16\\ (4/255)\end{tabular} & \begin{tabular}[c]{@{}c@{}}CW-100\\ (c=0.01)\end{tabular} & \begin{tabular}[c]{@{}c@{}}CW-100\\ (c=0.05)\end{tabular}    \\ \hline
$\sigma$=0.1    & 71.90                                                   & 32.54                                                   & \textbf{23.31}                                           & 79.58                                                     & 71.96                                                             \\ \hline
$\sigma$=0.01   & 71.56                                                   & 34.04                                                   & 25.96                                                    & 80.68                                                     & 72.00                                                               \\ \hline
$\sigma$=0.005  & 71.31                                                   & 27.79                                                   & 20.64                                                    & 80.46                                                     & 71.58                                                               \\ \hline
$\sigma$=0.001  & 70.51                                                   & 27.42                                                   & 17.98                                                    & 79.47                                                     & 67.00                                                               \\ \hline
\rowcolor[HTML]{EFEFEF} 
$\sigma$=0.05  & \textbf{74.18}                                          & \textbf{32.12}                                          & 22.12                                                    & \textbf{81.39}                                            & \textbf{74.72}                                             \\ \hline
\end{tabular}
}
\end{table}

\begin{table}[!h]
\centering
\caption{Accuracy (\%) on CIFAR-10 based on WRN-28-10 trained with the proposed method using various of $alpha$ for the Beta distribution to generate mixing coefficient $\lambda$. }
\label{tab:alpha}
\vspace{-5pt}
\setlength\tabcolsep{8pt}
\resizebox{.48\textwidth}{!}{
\begin{tabular}{c|ccccc}
\hline
Alpha          & \begin{tabular}[c]{@{}c@{}}FGSM \\ (8/255)\end{tabular} & \begin{tabular}[c]{@{}c@{}}PGD-8\\ (4/255)\end{tabular} & \begin{tabular}[c]{@{}c@{}}PGD-16\\ (4/255)\end{tabular} & \begin{tabular}[c]{@{}c@{}}CW-100\\ (c=0.01)\end{tabular} & \begin{tabular}[c]{@{}c@{}}CW-100\\ (c=0.05)\end{tabular}  \\ \hline
$\alpha$=2.0   & 71.27                                                   & 31.85                                                   & \textbf{22.39}                                           & 79.18                                                     & 70.50                                                             \\ \hline
\rowcolor[HTML]{EFEFEF} 
$\alpha$=1.0            & \textbf{74.18}                                          & \textbf{32.12}                                          & 22.12                                                    & \textbf{81.39}                                            & \textbf{74.72}                                            \\ \hline
\end{tabular}
}
\end{table}

\noindent
\textbf{Robust Image Recognition.}
To evaluate the adversarial robustness of the proposed method, we compare it with Mixup\cite{zhang2017mixup} and Manifold Mixup \cite{verma2019manifold} (denoted as M.-Mixup) and present the results in Table \ref{tab:ablation study} and Table \ref{tab:adv_cifar100_tiny}. 
As listed in Table \ref{tab:ablation study}, the average classification accuracy of the proposed method can reach $56.91\%$ on CIFAR-10, which surpasses Mixup and Manifold Mixup by margins of $37.31\%$ and $30.34\%$, respectively. 
Furthermore, the proposed model is also tested on a large-scale dataset, \emph{i.e.,} Tiny-ImageNet. As listed in Table \ref{tab:adv_cifar100_tiny}, the proposed approach can achieve superior performance against different kinds of adversarial attacks, compared to the other interpolation-based methods. Concretely, by using our AdaptiveMix, the model can achieve an average accuracy of $17.38\%$, surpassing Manifold Mixup by an improvement of $\sim$7\%.

Here, we analyze the influence caused by different values of hyper-parameters, including $\sigma$ for the noise term, and $\alpha$ for Beta Distribution in AdaptiveMix. As listed in Table \ref{tab:noise}, noises on multiple levels $\sigma$=[0.1, 0.05, 0.01, 0.005, 0.001] are considered for the grid search. The best performance is achieved as $\sigma$=0.05. 
In terms of $\alpha$, we conduct two settings for comparison, including $\alpha$=1.0 and 2.0 in Table \ref{tab:alpha}. The model trained with $\alpha$=1.0 is observed to outperform the one with $\alpha$=2.0 (\emph{i.e.,} $56.91\%$ \emph{vs.} $55.04\%$).

\begin{table}[]
\caption{Accuracy (\%) of the proposed AdaptiveMix on varying baselines and datasets. Res. stands for resolution of the input.}
\label{tab:cleanacc}
\vspace{-5pt}
\resizebox{.48\textwidth}{!}{
\begin{tabular}{c|c|c|c|
>{\columncolor[HTML]{EFEFEF}}c }
\hline
Dataset       & Architecture    & Res. & Baseline & Ours           \\ \hline
CIFAR-10      & WRN-28-10 \cite{zagoruyko2016wide}     & 32$^2$ & 96.11    & \textbf{96.80} \\ \hline
CIFAR-100     & WRN-28-10 \cite{zagoruyko2016wide}     & 32$^2$& 80.82    & \textbf{82.02} \\ \hline
T-ImageNet & PreActResNet-18 \cite{he2016identity}  & 64$^2$ & 57.23    & \textbf{60.59} \\ \hline
ImageNet      & ResNet-50 \cite{he2016deep}      & 128$^2$ & 67.38        & \textbf{68.69}     \\ \hline
\end{tabular}
}
\vspace{-10pt}
\end{table}

\noindent
\textbf{Clean Image Recognition.}
To validate the effectiveness of the proposed method on image recognition, we test the proposed method on various standard datasets and compare the results compared with the baseline \cite{zagoruyko2016wide,he2016identity,he2016deep}.  Table \ref{tab:cleanacc} shows  our method  improves the baseline on all the cases. In particular, on Tiny-ImageNet,  3\% absolute improvement can be achieved by the proposed method. The experimental result indicates that AdaptiveMix also not only improves the robustness but also benefits the generalization.

\begin{table}[tbp]
\caption{OOD detection on various OOD sets, where TIN-C, TIN-R, LSUN-C and LSUN-R refer to the OOD set of Tiny ImageNet-Crop, Tiny ImageNet-Resize, LSUN-Crop and LSUN-Resize, respectively. All values are F1 score ($\uparrow$), $\dagger$ stands for the result reproduced by the open-source code.  }
\label{tab:OOD}
\Huge
\vspace{-5pt}
\resizebox{.49\textwidth}{!}{
\begin{tabular}{ccccc}
\hline
\multicolumn{1}{c|}{ID Dataset}       & \multicolumn{4}{c}{CIFAR10}                                       \\ \hline
\multicolumn{1}{c|}{OOD Dataset}                  & TIN-C          & TIN-R          & LSUN-C         & LSUN-R         \\ \hline
\multicolumn{5}{c}{Methods using MC sampling}                                                                          \\ \hline
\multicolumn{1}{c|}{1DS \cite{zaeemzadeh2021out} (CVPR'21) }        & 0.930          & 0.936          & 0.962          & 0.961          \\ \hline\hline
\multicolumn{5}{c}{Methods which adopt OOD samples for validation and fine-tuning}                                   \\ \hline
\multicolumn{1}{c|}{ODIN \cite{liang2017enhancing} (ICLR'18) }                         & 0.902          & 0.926          & 0.894          & 0.937          \\
\multicolumn{1}{c|}{Mahalanobis  \cite{lee2018simple} (NIPS'18)}                  & 0.985          & 0.969          & 0.985          & 0.975          \\ \hline \hline

\multicolumn{1}{c|}{Soft. Pred. \cite{hendrycks2016baseline} (ICLR'17)}                & 0.803          & 0.807          & 0.794          & 0.815          \\
\multicolumn{1}{c|}{Counterfactual \cite{neal2018open} (ECCV'18)}               & 0.636          & 0.635          & 0.650          & 0.648          \\
\multicolumn{1}{c|}{CROSR \cite{yoshihashi2019classification} (CVPR'19)}                        & 0.733          & 0.763          & 0.714          & 0.731          \\
\multicolumn{1}{c|}{OLTR \cite{liu2019large} (CVPR'19)}                         & 0.860          & 0.852          & 0.877          & 0.877          \\ \hline
\multicolumn{1}{c|}{1DS  w/o MC $\dagger$ \cite{zaeemzadeh2021out}}        & 0.890          & 0.886          & 0.897          & 0.907          \\ 
\rowcolor[HTML]{EFEFEF} 
\multicolumn{1}{c|}{\cellcolor[HTML]{EFEFEF} 1DS  w/o MC $\dagger$ +Ours} & \textbf{0.922} & \textbf{0.911} & \textbf{0.934} & \textbf{0.937} \\ \hline

\end{tabular}
}
\vspace{-11pt}
\end{table}

\noindent
\textbf{OOD Detection.}
To validate the effectiveness of our method in OOD detection, we compare  with state-of-the-art OOD detection appoaches \cite{neal2018open,yoshihashi2019classification,liu2019large,zaeemzadeh2021out} on various datasets.  
We refer to  the accuracy  of 1DS  \cite{zaeemzadeh2021out}  as the  upper bound of other methods in Table \ref{tab:OOD},  since  1DS   employs Monte Carlo (MC) sampling which  sacrifices computational efficiency for achieving high accuracy.
Consequently, 1DS consumes  a much higher time cost than other methods.  
We  build  a baseline named "1DS  w/o MC"  \cite{zaeemzadeh2021out} removing  MC sampling for 1DS, and our method  combines "1DS  w/o MC"  with our AdaptiveMix.   Table \ref{tab:OOD} shows the performance of "1DS  w/o MC" is degraded due to the lack of   MC sampling. However,  our AdaptiveMix effectively improves  "1DS  w/o MC" without  expensive computational cost.

\section{Conclusion}
In this paper, we proposed a novel module named AdaptiveMix which is simple yet effectively  improves the training of GANs. 
By reducing the distance between training samples and their linear combination in a dynamic manner, AdaptiveMix can shrink regions of  training data in the feature  space, enabling the stable training of GANs and improving the image quality of generated samples.  We also demonstrate that AdaptiveMix is a reasonable way to ensure the approximate estimation of Lipschitz continuity.
Besides image generation, we show that AdaptiveMix can be applied to other tasks such as image classification and OOD detection, thanks to its plug-
and-play property.
Experimental results  demonstrate that our method effectively improves  the performance of  baseline models on seven publicly available datasets with regard to various tasks.

\section*{Acknowledgement}
This work was supported by the King Abdullah University of Science and Technology (KAUST) Office of Sponsored Research through the Visual Computing Center (VCC) funding, the Key-Area Research and Development Program of Guangdong Province, China (No. 2018B010111001), National Key R\&D Program of China (2018YFC2000702) and the Scientific and Technical Innovation 2030-"New Generation Artificial Intelligence" Project (No. 2020AAA0104100).  
{\small
\bibliographystyle{ieee_fullname}
\bibliography{cite}
}

\end{document}